\documentclass{llncs}
\usepackage{makeidx,nth}  
\usepackage{url}
\usepackage{graphicx,tabularx,diagbox}
\usepackage{booktabs,colortbl}
\usepackage{subfig}
\usepackage[group-separator={,}]{siunitx}
\usepackage[table]{xcolor}
\usepackage{pgfplots}
\usepackage[]{nth}

\usepgfplotslibrary{groupplots}
\pgfplotsset{compat=newest}
\newcolumntype{a}{>{\columncolor[gray]{0.9}}c}
\newcolumntype{b}{>{\columncolor[gray]{0.8}}c}

\begin{document}
\title{Spherical Harmonic Residual Network for Diffusion Signal Harmonization}
\titlerunning{SHResNet for Diffusion Signal Harmonization}

\author{Simon Koppers$^{1,2}$ \and Luke Bloy$^2$ \and Jeffrey I. Berman$^2$ \and Chantal M.W. Tax$^3$ \and J. Christopher Edgar$^2$ \and Dorit Merhof$^1$}
\institute{$^1$Institute of Imaging \& Computer Vision, RWTH Aachen University, Germany\\
$^2$Department of Radiology, Children's Hospital of Philadelphia, USA\\
$^3$CURBRIC, Cardiff University, UK}

\maketitle              

\begin{abstract}
Diffusion imaging\index{Diffusion Imaging} is an important method in the field of neuroscience, as it is sensitive to changes within the tissue microstructure of the human brain. 
However, a major challenge when using MRI to derive quantitative measures is that the use of different scanners, as used in multi-site group studies\index{multi-site study}, introduces measurement variability.
This can lead to an increased variance in quantitative metrics, even if the same brain is scanned. 

Contrary to the assumption that these characteristics are comparable and similar, small changes in these values are observed in many clinical studies, hence harmonization\index{Harmonization} of the signals is essential.

In this paper, we present a method that does not require additional preprocessing, such as segmentation or registration, and harmonizes the signal based on a deep learning\index{Deep Learning}residual network. 
For this purpose, a training database is required, which consist of the same subjects, scanned on different scanners.

The results show that harmonized signals are significantly more similar to the ground truth signal compared to no harmonization\index{Harmonization}, but also improve in comparison to another deep learning\index{Deep Learning} method.
The same effect is also demonstrated in commonly used metrics derived from the diffusion MRI signal.
\keywords{Harmonization, Diffusion Imaging, Magnetic Resonance Imaging, Machine Learning, Deep Learning, SHResNet}
\end{abstract}
\section{Introduction}
Diffusion imaging\index{Diffusion Imaging}, with diffusion tensor imaging (DTI) being the most commonly used strategy to represent the signal, is quickly becoming a critical non-invasive tool in the fields of neuroscience and neuropsychology.
Based on magnetic resonance imaging (MRI), diffusion imaging\index{Diffusion Imaging} is able to provide important insight into the structure of the human brain in both healthy- and patient populations.
Its ability to provide quantitative measurements sensitive to tissue microstructure, such as fractional anisotropy (FA) or mean diffusivity (MD), has lead to its widespread adoption in imaging research protocols.
With the recent increase in multi-site imaging studies\index{multi-site study}, it has become apparent that small differences in acquisition protocols across sites, but also differences of the scanner's vendor, can lead to significant differences in diffusion measurements even within the same subject.
Diffusion measurements between scanners can vary in the range of 5\% for white matter and 15\% for grey matter, potentially overwhelming the $\le$5\% effect sizes commonly found in group studies~\cite{Vollmar2010} and greatly offsetting the increases in statistical power that multi-site studies\index{multi-site study} and larger sample sizes can provide.

Data harmonization\index{Harmonization} is a collection of techniques designed to minimize inter-scanner and intra-site variance and to improve comparability of data acquired at different sites.
Currently, there are two strategies for harmonizing diffusion datasets.
The first uses a statistical normalization to create a common feature space~\cite{Fortin2017}.
Data from each site is normalized using site-specific means and variances.
A drawback is that each derived metric~(FA, MD, etc.) must be treated separately.
Furthermore, this approach doesn't allow the definition of a characteristic range of metrics, which would be important for comparing single subjects or other multi-site studies\index{multi-site study}.

The second harmonization\index{Harmonization} approach is to learn a mapping between the data of all the sites and a pre-defined standard system~\cite{Mirzaalian2015}.
In this case, the measured signals, and therefore all derived diffusion metrics, are harmonized and made comparable.
This allows the same analysis pipeline to be subsequently applied to all data records regardless of acquisition site.
Currently, there are only few published methods~\cite{Mirzaalian2015,Mirzaalian2018}, which are model-independent methods that harmonizes diffusion data based on a template-based mapping of rotation invariant spherical\index{spherical} harmonic (RISH) features.
RISH features are first calculated for every subject and used to create a single template mapping for each RISH order.
Afterwards, every subject can be corrected based on the corresponding RISH template mapping, which is a rotation invariant mapping, maintaining the original fiber orientations.
However, the disadvantage of this method is that an incorrect registration during the template generation, but also during application, leads to an unreliable correction template.
In addition, morphological changes, such as tumor resections or edemas, might lead to additional registration artifacts.

While not developed explicitly for harmonization\index{Harmonization}, a number of recently developed deep learning\index{Deep Learning} techniques can potentially be adapted for this purpose.
In general, deep learning\index{Deep Learning} methods~\cite{Golkov2015} can be used to build an internal predictive model of diffusion signals, and have shown good results in interpolating diffusion signals, but also predicting complete shells based on only a few gradient directions~\cite{Koppers2016}.

In this work, we propose a novel customized deep learning\index{Deep Learning} method based on a residual structure to harmonize diffusion data from different systems, which only has to learn the difference between two signals.
In addition, it is independent of registration during application, which is why morphological changes have only minor influences on its harmonization\index{Harmonization} performance.
In this case the full spherical\index{spherical} signal based on spherical\index{spherical} harmonics (SH) is predicted, while for multi-shell data other general bases are possible.
%
%
%
\section{Method}
The goal of diffusion signal harmonization\index{Harmonization} is to increase the comparability of diffusion signals measured on two different MRI systems.
Our approach is to use a spherical\index{spherical} harmonic residual network to predict the data acquired on the second scanner using the data from the first scanner.
Once the network is trained on a small number of subjects acquired on both scanners, it can be applied to new datasets acquired on the \nth{1} scanner and outputs diffusion data that matched the properties of the second scanner.
In addition, a RISH projection as in~\cite{Mirzaalian2017} is performed on every resulting harmonized signal in order to enforce the same fiber orientations as in the input signal.

\subsection{Spherical Harmonic Residual Network}
The Spherical\index{spherical} Harmonic Residual Network (SHResNet\index{SHResNet}) can be described as a method that takes a $3\times 3\times3$ spherical\index{spherical} harmonic input patch, acquired on scanner 1, and predicts the spherical\index{spherical} harmonic coefficients of the central voxel, within the same patch, in scanner 2.
It is based on a residual structure~\cite{He2016}, which is generally known to be very robust due to their forward pass and are therefore able to train very deep network structures without losing the ability to be trained based on the vanishing gradient problem~\cite{Bengio1994}.
\begin{figure*}[!htb]
	\centering
		\includegraphics[width=1\textwidth]{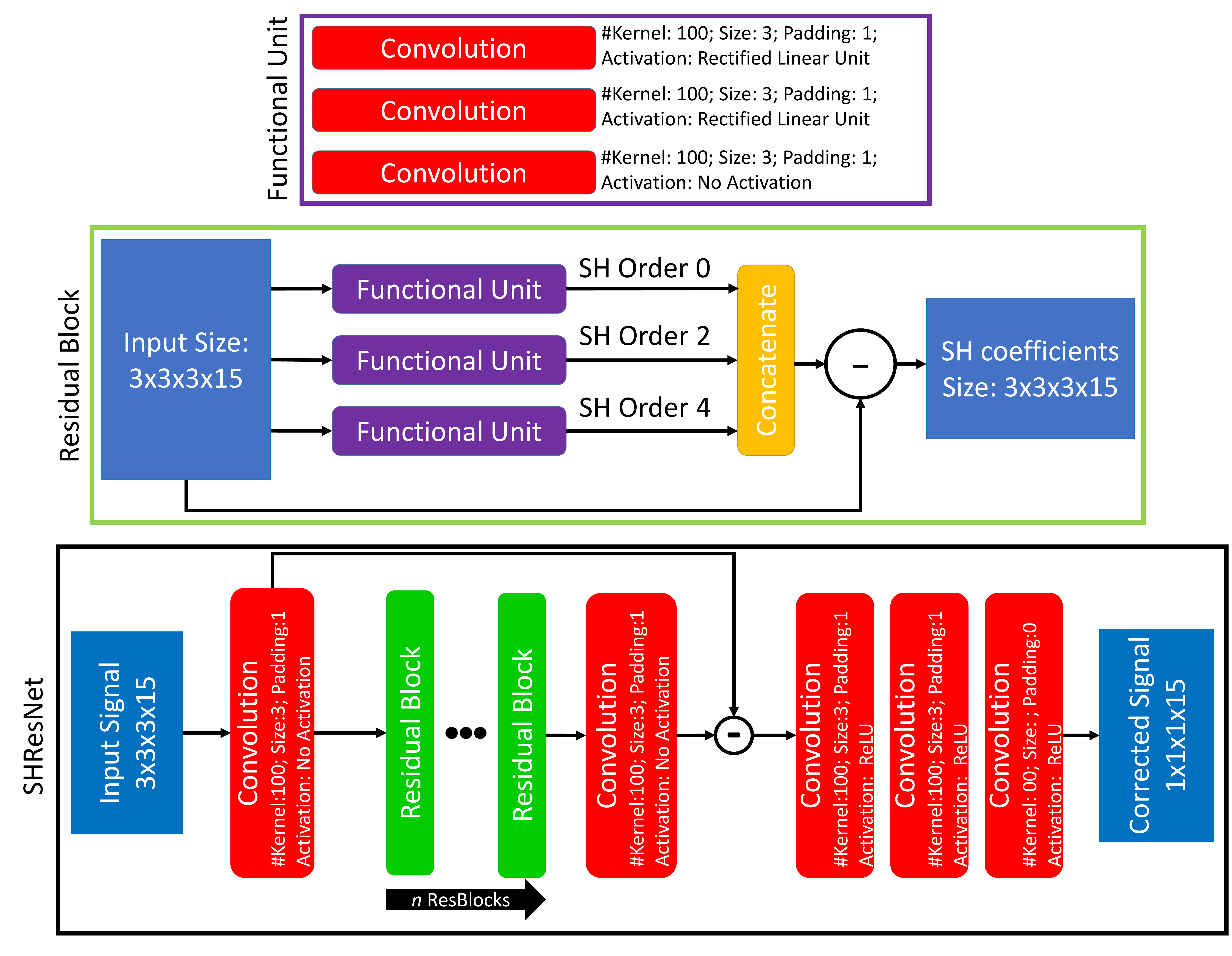}
	\caption{Structure of the complete SHRestNet.}
	\label{fig:net}
\end{figure*}
The network itself can be separated into three parts as shown in Fig.~\ref{fig:net}.
The core structure is the residual block (ResBlock), which can be stacked multiple times, keeping spatial and coefficient dimensions.
Its main element is called functional unit, which consists of three 3D convolutional layers (kernel size: $3\times 3\times3$ with padding) predicting a single SH order.
Depending on the corresponding SH order, the number of output neurons needs to be adjusted (SH order 0: 1 neuron; SH order 2: 5 neurons; SH order 4: 9 neurons).
The first two convolutional layers utilize a rectified linear unit (ReLU) as activation function.
The resulting coefficients are concatenated and subtracted from the functional unit's input.
The full network uses two 3D convolutional layers (kernel size: $3\times 3\times3$ with padding) as preprocessing and postprocessing layers, with multiple ResBlocks in the middle.
After that, an additional residual pathway subtracts the preprocessed input from the postprocessed output of all ResBlocks.
At the final stage, 3D convolutional layers, employing spatial neighboring information, are utilized to reduce the input dimensions from $3\times 3\times 3$ to the single central voxel.

After each signal is harmonized in SH space, it is corrected for possible changes in fiber orientation.
For this purpose, each signal's RISH features are projected onto the corresponding input signal based on
\begin{equation}
S_{i}' = S_{i} \cdot \frac{\sqrt{RISH_{i}'}}{\sqrt{RISH_i}} \mathrm{,}
\label{eq:rish}
\end{equation}
where $S_{i}'$ describes the harmonized and $S_i$ the non-harmonized spherical\index{spherical} harmonic coefficients of order $i$, while $RISH_i'$ and $RISH_i$ describe the RISH feature of the harmonized and the input signal, respectively.
In the end, the signal space is reconstructed by sampling each gradient direction utilizing the predicted SH coefficients, while every predicted gradient is multiplied by the non-diffusion weighted signal ($b=0\frac{\mathrm{s}}{\mathrm{mm^2}}$) of the initial measurement in the input system.

\subsubsection{Training}
The network is trained on a sample of subjects scanned on both systems.
For each training subject, diffusion data from the first scanner acquisition is spatially registered to the target, and training voxels (grey and white matter) are identified using FAST~\cite{Zhang2001}.
Training is performed by looping over each training subject, attempting to predict the second scanner's data using the diffusion data from the first.
The network is trained in 2 stages, the first utilizing the Adam optimizer (learning rate of 0.001, batch size: 256 3D samples), which is replaced, in the \nth{2} stage after the first five epochs, by an SGD optimizer.
During the \nth{2} stage the batch size is reduced to 128 samples and the learning rate decreases by ten percent if the validation performance does not improve over five epochs.
Training is completed after validation performance does not improve for more than ten epochs.
Overall, the mean squared error (MSE) in signal space is utilized as loss function.
Furthermore, RISH projection is not part of the training cycle and was only applied during the application stage.
\section{Results}
We present the results for two experiments, the first~(Sec.~\ref{sec:eva_res}) investigates the sensitivity of the method to the number of ResBlocks $n$ required to achieve optimal harmonization\index{Harmonization}.
In the second section~(Sec.~\ref{sec:eva_harm}), we evaluate our methods ability to reduce inter-scanner variance.
As our method harmonizes the diffusion signal directly it subsequently harmonizes any DTI marker computed from the diffusion signal.
For the purposes of our comparison we investigate differences in FA, MD~\cite{Jenkinson2012} and the signal itself between our method (SHResNet\index{SHResNet}), an adapted version of the network of Golkov et al.~\cite{Golkov2015}, hereinafter referred to as the Golkov method, and non-harmonized values.
The Golkov method utilizes a three-layer neural network and is usually used to interpolate diffusion signals, but also to reconstruct specific diffusion quantities.
It's output is adapted i.e. it is able to predict SH coefficients.
%
\subsection{Material}
For evaluating the proposed method, 10 uncorrelated healthy subjects were selected from the CDMRI Harmonization\index{Harmonization} Challenge~\cite{CDMRIChallenge}\index{challenge} database, while grey and white matter~\cite{Zhang2001} are evaluated separately to have a more detail insight into the method, while a segmentation is not utilized during training..
Each subject was scanned on both a 3T GE Excite-HD and a 3T Siemens Prisma scanner utilizing the same protocol.
Both scans have a isotropic resolution of 2.4 mm$^3$, and 30 gradient directions at $b = 1200 \frac{\mathrm{s}}{\mathrm{mm}^2}$, as well as a minimum of 4 baseline measurements with $b = 0 \frac{\mathrm{s}}{\mathrm{mm}^2}$.
All data were corrected for motion and eddy current-distortions~\cite{Andersson2003}, while the Prisma data was corrected for susceptibility distortions~\cite{Andersson2016}.
Afterwards, the GE dataset was spatially nonlinearly registered to the corresponding Prisma dataset~\cite{Klein2010}.

In order to be applicable to every gradient setting, each dataset was preprocessed by dividing every voxel's diffusion signal by its non-diffusion measurement ($b=0\frac{\mathrm{s}}{\mathrm{mm^2}}$) and further converted utilizing spherical\index{spherical} harmonics (SH) of order four, limited due to the amount of gradient directions, and a regularization of $\lambda=0.006$, resulting in 15 SH coefficients.
\subsection{Evaluation of number of ResBlocks}
\label{sec:eva_res}
To estimate the optimal number of ResBlocks $n$, harmonization\index{Harmonization} is evaluated utilizing the mean cross validation loss over 10 subjects.
As shown in Tab.~\ref{tab:n}, the SHResNet\index{SHResNet} reaches its most stable accurate performance utilizing $n=2$ ResBlocks within the network, while higher $n$ result to a high error.
\begin{table*}
\centering
\caption{Resulting MSE Loss (scaled by 100) for different number of ResBlocks during optimization.}
\begin{tabular}{lccccccc} 
\toprule
Loss & 1 & 2 & 3 & 6 & 7 & 9 & 11\\ 
\midrule 
MSE & 0.375 & 0.372 & 0.375 & 0.377 & 0.41 & 0.41 & 0.41\\ 
\bottomrule
\end{tabular}
\label{tab:n}
\end{table*}
In addition, it can be seen that $n=1$ ResBlocks already results in a similar MSE compared to $n=2$ ResBlocks.

\subsection{Evaluation of Harmonization}
\label{sec:eva_harm}
In order to evaluate our method we preformed a 10 fold cross validation.
For each fold a SHResNet\index{SHResNet} ($n=2$ ResBlocks) was trained on 8 subjects, with 1 validation subject, using all grey and white matter voxels, to predict the Prisma diffusion data using the GE data as input.
Afterwards, the trained SHResNet\index{SHResNet} was used to predict the Prisma diffusion signal of the test subject.
For comparison the Golkov method was also used to generate a predicted Prisma diffusion subject using the same training strategy.
After looping over each fold, for each subject, there are 3 predicted Prisma datasets (SHResNet\index{SHResNet}, Golkov and the unharmonized GE dataset) that were then compared to the ground truth Prisma diffusion dataset for that subject.

We utilized the normalized mean squared error (NMSE)(see Eq.~\ref{eq:nmse}) of the diffusion signal its resulting FA and MD, to contrast the four methods.
\begin{equation}
\mathrm{NMSE}=\frac{\left\|\mathrm{Y} - \hat{\mathrm{Y}}\right\|_{2}^2}{\left\|\mathrm{Y}\right\|_{2}^2} \mathrm{,}
\label{eq:nmse}
\end{equation}
where $\mathrm{Y}$ represents the ground truth vector and $\hat{\mathrm{Y}}$ is the predicted vector.
Furthermore, a pair-wise Wilcoxon signed rank test was performed on the resulting differences to evaluate the statistical improvement.

\begin{figure}[ht]
     \subfloat[White matter\label{sig:wm_signal}]{%
       \includegraphics[width=0.48\textwidth]{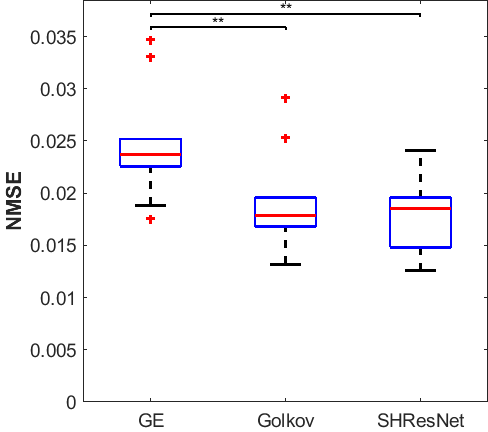}
     }
  \hspace{\fill} 
     \subfloat[Grey matter\label{sig:gm_signal}]{%
       \includegraphics[width=0.48\textwidth]{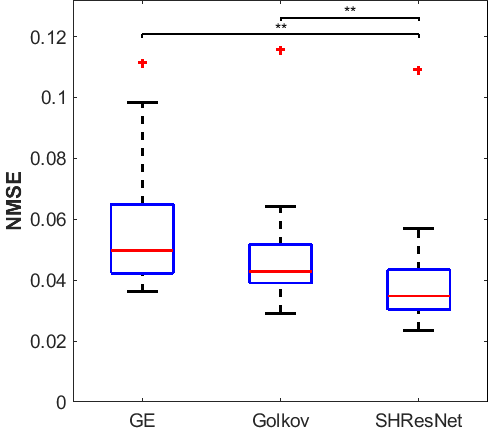}
     }
\caption{Resulting NMSE for the signal in white and grey matter ($\ast: p\le 0.05$ and $\ast\ast: p\le 0.01$).}
\label{fig:block_signal} 
\end{figure}

\begin{figure}[ht]
     \subfloat[White matter\label{sig:wm_fa}]{%
       \includegraphics[width=0.48\textwidth]{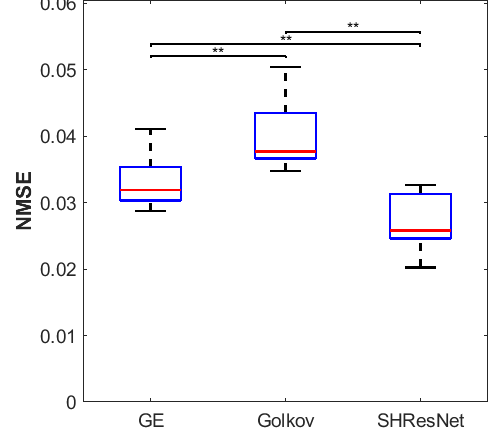}
     }
  \hspace{\fill} 
     \subfloat[Grey matter\label{sig:gm_fa}]{%
       \includegraphics[width=0.48\textwidth]{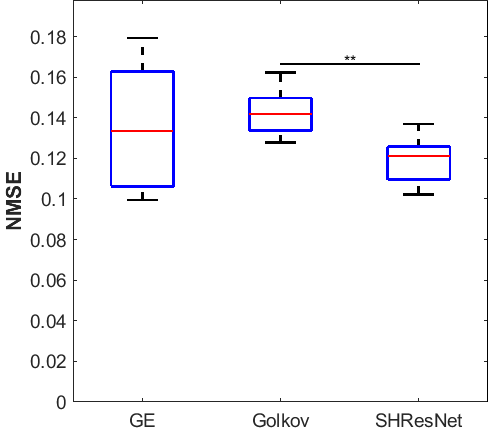}
     }
\caption{Resulting NMSE for the FA in white and grey matter ($\ast: p\le 0.05$ and $\ast\ast: p\le 0.01$).}
\label{fig:block_fa} 
\end{figure}

\begin{figure}[ht]
     \subfloat[White matter\label{sig:wm_md}]{%
       \includegraphics[width=0.48\textwidth]{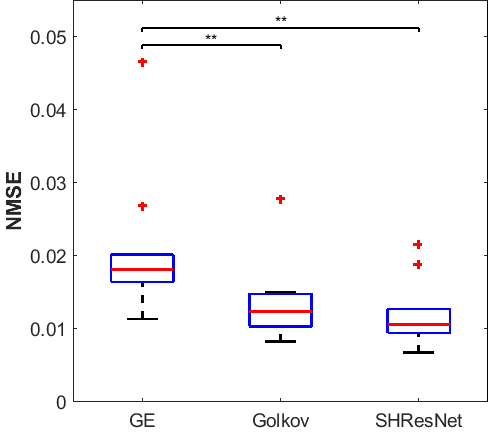}
     }
  \hspace{\fill} 
     \subfloat[Grey matter\label{sig:gm_md}]{%
       \includegraphics[width=0.48\textwidth]{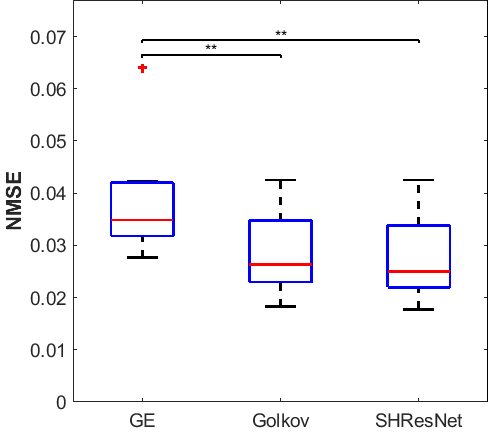}
     }
\caption{Resulting NMSE for the MD in white and grey matter ($\ast: p\le 0.05$ and $\ast\ast: p\le 0.01$).}
\label{fig:block_md} 
\end{figure}

Fig.~\ref{fig:block_signal}, Fig.~\ref{fig:block_fa} and Fig.~\ref{fig:block_md} present the resulting NMSE comparisons for white and grey matter voxels respectively.
It can be seen that the SHResNet\index{SHResNet} achieves a significantly lower NMSE in every scenario for white matter and a significant lower error in grey matter for the signal and the MD, while the improvement in FA only results in a non-significant p-value of $\approx$0.08.
Further, it should be noted that the Golkov method improves the harmonization\index{Harmonization} accuracy for the signal, the SH coefficients and the MD, while it results in a decreased performance for the FA in white and grey matter.

On average, the signal error decreased by $\approx$35$\%$ in white matter and $\approx$29$\%$ in grey matter utilizing the SHResNet\index{SHResNet}.
A similar improvement can also be seen for the FA (white: $\approx$18$\%$; grey:$\approx$9$\%$) and the MD (white: $\approx$40$\%$; grey:$\approx$28$\%$).

\section{Discussion}
Diffusion acquisitions are increasingly be included in large multi-site studies\index{multi-site study}.
While the large sample sizes afforded by these types of studies offer great promise in illucidating population differences, the high levels of inter-site variability, particularly in diffusion but also in other imaging modalities, greatly diminishes the ability to observe small effects.
Data harmonization\index{Harmonization} represents a post processing technique designed to minimize inter-site variability.
In this paper, we present a novel diffusion data harmonization\index{Harmonization} technique, the SHResNet\index{SHResNet}, based on deep learning\index{Deep Learning}, and demonstrate its capability to reduce inter-site and inter-scanner variance.

As shown in Tab.~\ref{tab:n}, training on deeper networks (increased number of ResBlocks) does not improve harmonization\index{Harmonization}.
This effect can be due to several reasons.
First, deeper networks are generally more difficult to train, because of the significantly increased number of parameters, while at the same time each voxel also contains a complete diffusion signal.
Furthermore, the input size was limited to $3\times 3\times 3$ voxels.
Deeper networks, however, are primarily intended for larger region-specific neighborhood relationships, which are not desirable in our case, since it could lead to problems in scenarios of anatomical changes.

The evaluation of our method (see Fig.~\ref{fig:block_signal}, Fig.~\ref{fig:block_fa} and Fig.~\ref{fig:block_md}) shows that it achieves a significant improvement compared to non-harmonized data, while the Golkov method seem to have difficulties reconstructing the correct signal in white matter, resulting in a higher FA NMSE in comparison to the non-harmonized case.

Similar, while not always significant, effects can be seen in the area of grey matter, even though the underlying microstructure is much more complex as in white matter.

Moreover, we also participated with a simpler version of this model (SHResNet\index{SHResNet} with $n=1$ without RISH projection) in the CDMRI Harmonization\index{Harmonization} Challenge\index{challenge} 2017 and consistently showed a very stable and good performance in all given tasks~\cite{CDMRIChallenge}.

Despite these promising results, there are also some limitations.
The biggest obstacle of this algorithm and harmonization\index{Harmonization} in general is the applicability in a clinical study environment.
Since it is only able to reconstruct scenarios represented by the utilized training set, unique features occurring in clinical cases may not be preserved through the harmonization\index{Harmonization} process.
Thus for population studies it is important that the training sample, that is scanned on both systems, have similar symptomatology and demographic distributions as the entire study sample.
Using synthetic data that has similar diffusion characteristics as found in the clinical subjects, the method may be better able to handle these use cases.
These types of comparisons are beyond the scope of this work, but represent the next steps needed to use data harmonization\index{Harmonization} in large studies.
Furthermore, this method is not free of any registration, since a inter-subject registration is required between scanners to generate the training dataset.
\section{Conclusion}
Overall, the SHResNet\index{SHResNet} has been shown to be a very fast, uncomplicated and uniform approach to harmonize diffusion data.
It is validated utilizing the NMSE with a significant improvement after harmonization\index{Harmonization} found for the signal, the FA and the MD in white and grey matter.
\section{Acknowledgments}
The authors would like to thank the 2017 computational dMRI challenge organizers (Francesco Grussu, Enrico Kaden, Lipeng Ning and Jelle Veraart) for help with data acquisition and processing, as well as Derek Jones, Umesh Rudrapatna, John Evans, Slawomir Kusmia, Cyril Charron, and David Linden at CUBRIC, Cardiff University, and Fabrizio Fasano at Siemens for their support with data acquisition. This work was supported by a Rubicon grant from the NWO (680-50-1527), a Wellcome Trust Investigator Award (096646/Z/11/Z), and a Wellcome Trust Strategic Award (104943/Z/14/Z). The data were acquired at the UK National Facility for In Vivo MR Imaging of Human Tissue Microstructure funded by the EPSRC (grant EP/M029778/1), and The Wolfson Foundation.

This work was supported by the International Research Training Group 2150 of the German Research Foundation (DFG).
%
%
%
\bibliographystyle{splncs03}
\bibliography{koppers}
\end{document}